\documentclass{ecai}
\usepackage{times}
\usepackage{graphicx}
\usepackage{latexsym}

\usepackage{amsmath,amssymb,amsthm}
\usepackage{booktabs}
\usepackage{algorithm}

\usepackage{eurosym}
\usepackage{url}
\usepackage{subfig}
\usepackage{multirow}
\usepackage{algpseudocode}
\usepackage{xspace}
\usepackage{xcolor}
\usepackage{xargs} 
\usepackage[colorinlistoftodos,prependcaption,textsize=tiny]{todonotes}
\newcommandx{\add}[2][1=]{\todo[linecolor=red,backgroundcolor=red!25,bordercolor=red,#1]{#2}}
\newcommandx{\change}[2][1=]{\todo[linecolor=blue,backgroundcolor=blue!25,bordercolor=blue,#1]{#2}}
\newcommandx{\info}[2][1=]{\todo[linecolor=OliveGreen,backgroundcolor=OliveGreen!25,bordercolor=OliveGreen,#1]{#2}}
\newcommandx{\improvement}[2][1=]{\todo[linecolor=orange,backgroundcolor=orange!25,bordercolor=orange,#1]{#2}}

\newtheorem{example}[theorem]{Example}
\theoremstyle{definition}
\newtheorem{definition}{Definition}[section]

\newcommand{\trepan}{\textsc{Trepan}\xspace}

\newcommand{\TREPAN}{\textsc{TREPAN}\xspace}

\newcommand{\Tmc}{\ensuremath{\mathcal{T}}\xspace}

\newcommand{\ELb}{\ensuremath{\mathcal{E\!L}_\bot}\xspace}

\newcommand{\onto}[1]{\ensuremath{\mathsf{#1}}}

\newcommand{\sub}{\ensuremath{\mathsf{sub}}\xspace}

\ecaisubmission   

\begin{document}

\title{\trepan Reloaded: A Knowledge-driven Approach to Explaining Black-box Models}

\author{Roberto~Confalonieri\institute{Telef{\'o}nica Innovaci{\'o}n Alpha, email:  \url{{roberto.confalonieri,tarek.besold,fermin.moscoso}@telefonica.com}}  \and Tillman~Weyde\institute{Dept. of Computer Science, City, University of London, email: \texttt{t.e.weyde@city.ac.uk}} \and Tarek~R.~Besold$^1$
\and Ferm{\'in}~Moscoso~del~Prado~Mart{\'i}n$^1$}

\maketitle
\bibliographystyle{ecai}

\begin{abstract}
Explainability in Artificial Intelligence has been revived as a topic of active research by the need of conveying safety and trust to users in the `how' and `why' of automated decision-making. 
Whilst a plethora of approaches have been developed for post-hoc explainability, only a few focus on how to use domain knowledge, and how this influences the understandability of global explanations from the users' perspective. 
In this paper, we show how ontologies help 
the understandability of global post-hoc explanations, 
presented in the form of symbolic models. 
In particular, we build on \trepan, an algorithm that explains artificial neural networks by means of decision trees, and we extend it to include ontologies modeling domain knowledge in the process of generating explanations.
We present the results of a user study that measures the understandability of decision trees using a syntactic complexity measure, and through time and accuracy of responses as well as reported user confidence and understandability. The user study considers domains where explanations are critical, namely, in finance and medicine. 
The results show that decision trees generated with our algorithm, taking into account domain knowledge, are more understandable than those generated by standard \trepan without the use of ontologies.
\end{abstract}

\section{INTRODUCTION}

In recent years, explainability has been identified as a key factor for the adoption of AI systems in a wide range of 
contexts~\cite{Confalonieriatal2019cogsci,HoffmanMKL18,Doshi-Velez2017,Lipton2018,Ribeiro2016,Miller2017}.
The emergence of intelligent systems in self-driving cars, medical diagnosis, insurance and financial services among others has shown that when decisions are taken or suggested by automated systems it is essential for practical, social, and 
increasingly legal reasons that an explanation can be provided to users, developers or regulators. 
As a case in point, the European Union's General Data Protection Regulation (GDPR) stipulates a right to ``{\em meaningful information about the logic involved}''---commonly interpreted as a `right to an explanation'---for consumers affected by an automatic decision~\cite{GDPR2016}.\footnote{Regulation (EU) 2016/679 on the protection of natural persons with regard to the processing of personal data and on the free movement of such data, and repealing Directive 95/46/EC (General Data Protection Regulation) [2016] OJ L119/1.} 

The reasons for equipping intelligent systems with explanation capabilities are not limited to user rights and acceptance. Explainability is also needed for designers and developers to enhance system robustness and enable diagnostics to prevent bias, unfairness and discrimination~\cite{mehrabi2019survey}, as well as to increase trust by all users in {\em why} and {\em how} decisions are made. 
Against that backdrop, increasing efforts are directed towards studying and provisioning explainable intelligent systems, both in industry and academia, sparked by initiatives like the DARPA Explainable Artificial Intelligence Program (XAI), 
and carried by 
a growing number of scientific conferences and workshops dedicated to explainability. 

While interest in XAI had subsided 
together with that in expert systems after the mid-1980s~\cite{Buchanan1984,Wick1992}, %
recent successes in machine learning technology have brought explainability back into the focus. 
This has led to a plethora of new approaches for local and global {\em post-hoc} 
explanations of black-box models~\cite{Guidotti2018}, for both autonomous and  human-in-the-loop systems, aiming to achieve explainability without sacrificing system performance. 
Only a few of these approaches, however, focus on how to integrate and use domain knowledge to drive the explanation process (e.g.,~\cite{Towell1993,Renard2019ConceptTree}) 
or to measure the understandability of explanations of black-box models (e.g.,~\cite{Ribeiro2018}). For that reason an important foundational aspect of explainable AI remains hitherto mostly unexplored: Can the integration of domain knowledge as, e.g., modeled by means of ontologies, help the understandability of interpretable machine learning models?

To tackle this research question, we propose a neural-symbolic learning approach based on \trepan~\cite{Craven1995}, an algorithm devised in order to explain trained
artificial neural networks by means of decision trees, and we extend it to take into account ontologies in the explanation generation process. In particular, we modify the logic of the algorithm when choosing split nodes, to prefer features associated with more general concepts in a domain ontology. Having explanations bounded to structured knowledge, in the form of ontologies, conveys two advantages. First, 
it enriches explanations (or the elements therein) with semantic information, and facilitates effective knowledge transmission to users. 
Second, it supports the customisation of the levels of specificity and generality of explanations to specific user profiles~\cite{Hind2019exXAI}. 

In this paper, we focus on the first advantage, and on measuring the impact of the ontology on the perceived understandabilty of surrogate decision trees. To evaluate our approach, we designed and conducted an experiment
to measure the understandability of decision trees in domains where explanations are critical, namely the financial and medical domain. 
Our study shows that decision trees generated by our modified \trepan  algorithm taking domain knowledge into account are more understandable than those generated without the use of domain knowledge. Crucially, this enhanced understandability of the resulting trees is achieved with little compromise on the accuracy with which the resulting trees replicate the behaviour of the original neural network model.

The remainder of the paper is organised as follows. 
After introducing \trepan, and the notion of ontologies (Section~\ref{sec:preliminaries}), we present our revised version of the algorithm that takes into account ontologies in the decision tree extraction (Section~\ref{sec:trepan_reloaded}). In Section~\ref{sec:understandability}, we propose how to measure understandability of decision trees from a technical and a user perspective.  Section~\ref{sec:evaluation} reports and analyses the results of our experiment. After discussing our approach (Section~\ref{sec:discussion}), Section~\ref{sec:related_work} situates our results in the context of related contributions in XAI. Finally, Section~\ref{sec:conclusion} concludes the paper and outlines possible lines of future work.

\section{PRELIMINARIES}\label{sec:preliminaries}

In this section, we present the main foundations of our approach, namely, the \trepan algorithm and ontologies.

\subsection{The \trepan algorithm}

\trepan is a tree induction algorithm that recursively extracts decision trees from oracles, 
in particular from feed-forward neural networks~\cite{Craven1995}. 
The original motivation behind the development of \trepan was to approximate a neural network by means of a symbolic structure that is more interpretable than a neural network classification model. 
This was in the context of a wider interest in knowledge extraction from neural networks (see~\cite{Towell1993,AvilaGarcez2001} for an overview). 

\begin{algorithm}[t]
\begin{algorithmic}
\State Priority queue $Q \gets \emptyset$ 
\State Tree $T \gets \emptyset$
\State use $Oracle$ to label examples in $Training$ 
\State enqueue root node into $Q$
\While{$nr\_internal\_nodes < size\_limit$}
\State pop node $n$ from $Q$ 
\State generate $examples$ for $n$
\State use $Features$ to build set of candidate $splits$
\State use $Examples$ and $Oracle$ to determine $Best\_split$ 
\State add $n$ to $T$
\For{element $c \in Best\_split$}
\State add $c$ as child of $n$
\If {$c$ is not a leaf according to the $Oracle$}
\State enqueue node $c$ into $Q$ with negative 
\State \indent information gain as priority
\EndIf
\EndFor
\EndWhile
\State Return $T$
\end{algorithmic}
\caption{Trepan($Oracle$,$Training$,$Features$) . 
}
	\label{algo:trepan}
\end{algorithm}
The pseudo-code for \trepan is shown in Algorithm~\ref{algo:trepan}. 
\trepan differs from conventional inductive %
learning algorithms 
as it uses an {\em oracle} to classify examples during the learning process. 
It generates new examples by sampling from distributions over the given examples and constraints, so that the amount of training data used to select splitting tests and to label leaves does not decrease with the depth of the tree. 
It expands a tree in a best-first manner by means of a priority queue by entropy, that prioritises nodes that have greater potential for improvement. 
Further details of the algorithm can be found in~\cite{Craven1995}.

\trepan stops the tree extraction process using two criteria: all nodes 
do not need to be further expanded because their entropy is low (they contain
almost exclusively %
instances of a single class), or a predefined limit of the tree size (the number of nodes) is reached. 
Whilst \trepan was designed to explain neural networks as the oracle,
it is a model-agnostic algorithm and can be used to explain any other classification 
model.

In this paper, our objective is to improve the understandability of the decision trees extracted by \trepan. 
To this end, we extend the algorithm 
to take into account an {\em information content} measure, that is derived using ontologies, and computed using the idea of concept refinement, as detailed below. In order to evaluate the performance of both the original and extended \trepan algorithms, we measure the accuracy and the fidelity of the resulting decision trees. {\em Accuracy} is defined as the percentage of test-set examples that are correctly classified. In contrast, {\em fidelity} is defined as the percentage of test-set examples on which the classification made by a tree agrees with that provided by its neural-network counterpart. Notice that the crucial measure for assessing the quality of the reconstructed tree is the fidelity, as this is the direct measure of how well the tree's behaviour mimics the original neural network.

\begin{figure}[!t]
\resizebox{\columnwidth}{!}{
\fbox{
	\begin{minipage}[t]{12cm}
		\begin{tabular}{ll}
			\onto{Entity \sqsubseteq \top}  \qquad & , \qquad \onto{Person \sqsubseteq PhysicalObject}  \\
			
			\onto{AbstractObject \sqsubseteq Entity} \qquad & , \qquad \onto{Loan \sqsubseteq AbstractObject}  \\
			
			\onto{PhysicalObject \sqsubseteq Entity} \qquad & , \qquad  \onto{Gender  \sqsubseteq Quality}  \\
			
			\onto{Quality \sqsubseteq Entity} \qquad   & , \qquad \onto{Male \sqsubseteq Gender}  \\
            
            \onto{LoanApplicant  \sqsubseteq Person \sqcap \exists hasApplied.Loan} \qquad   & , \qquad \onto{Female \sqsubseteq Gender}  \\			
            
    
			
\onto{Domain(hasApplied) = Person} \qquad & , \qquad \onto{Range(hasApplied) = Loan}  \\

		\end{tabular}
\end{minipage}
}}
\caption{An ontology 
excerpt for the loan domain.}
\label{fig:ontology}
\end{figure}
 
\subsection{Ontologies}
\label{sec:ontologies}


An ontology is a set of 
formulae %
in an appropriate logical language
with the purpose of describing a particular domain of interest, 
such as 
finance or medicine.  
The precise logic used is not crucial for our approach as 
the techniques introduced here apply to a variety of logics. 
For the sake of clarity we use description logics (DLs) as well-known ontology languages. 
We briefly introduce the DL
\ELb, 
a DL allowing only conjunctions,  existential restrictions, and the empty concept $\bot$. 
For full details, see~\cite{BaaderDLH03,DBLP:conf/ijcai/BaaderBL05}. 
\ELb is widely used in biomedical ontologies for describing large terminologies and it is the base of the OWL 2 EL profile. 

Syntactically, \ELb is based on two disjoint sets $N_C$ and $N_R$ of \emph{concept names}
and \emph{role names}, respectively.
The set of \emph{\ELb concepts} is generated by the grammar 
\begin{eqnarray*}
C & ::= & A\mid C \sqcap C \mid \exists R.C \enspace,
\end{eqnarray*}
where $A\in N_C$ and $R\in N_R$. 
A \emph{TBox} is a finite set of general concept inclusions (GCIs) of the form $C\sqsubseteq D$ where $C$ and 
$D$ are concepts. It stores the terminological knowledge regarding the relationships between 
concepts. 
An \emph{ABox} is a finite set of assertions $C(a)$ and $R(a,b)$, which express 
knowledge about objects in the knowledge domain.
An \emph{ontology} is composed by a TBox and an ABox. In this paper, we focus on the TBox only, thus we will use the terms ontology and TBox interchangeably.

The semantics of \ELb is based on \emph{interpretations} of the form $I = (\Delta^I, \cdot^I)$, 
where $\Delta^I$ is a non-empty \emph{domain}, and $\cdot^I$ is a function mapping every
individual name to an element of $\Delta^I$, each concept name to a subset of the domain, and each role 
name to a binary relation on the domain. $I$ satisfies $C \sqsubseteq D$ iff $C^I \subseteq D^I$ and $I$ satisfies an assertion $C(a)$ ($R(a,b)$) iff 
$a^{I} \in C^{I}$ ($(a^{I},b^{I}) \in R^{I}$).
%
The interpretation $\mathcal{I}$ is a \emph{model} of the TBox \Tmc if it satisfies all the GCIs and 
all the assertions in \Tmc. \Tmc is \emph{consistent} if it has a model.
Given two concepts $C$ and $D$,  $C$ is \emph{subsumed} by $D$ w.r.t. \Tmc ($C \sqsubseteq_{\Tmc} D$) if $C^I \subseteq D^I$ for every model 
$I$ of \Tmc. We write $C \equiv_{\Tmc} D$ when $C \sqsubseteq_{\Tmc} D$ and $D \sqsubseteq_{\Tmc} C$.
$C$ is \emph{strictly subsumed by} $D$ w.r.t.\ \Tmc ($C \sqsubset_{\Tmc} D$) if 
$C \sqsubseteq_{\Tmc} D$ and $C \not\equiv_{\Tmc} D$.


Figure~\ref{fig:ontology} shows an ontology excerpt modeling concepts and relations relevant to the {\em loan} domain. The precise formalisation of the domain is not crucial at this point; different formalisations may exist, with different levels of granularity.
The ontology 
structures the domain knowledge from the most {\em general} concept (e.g.,~\onto{Entity}) to more {\em specific} concepts (e.g.,~\onto{LoanApplicant}, \onto{Female}, etc.). The subsumption relation ($\sqsubseteq$) induces a partial order among the concepts that can be built from a TBox \Tmc. 
For instance, the \onto{Quality} concept is more general than the \onto{Gender} concept, and it is more specific than the \onto{Entity} concept. 

We will capture the degree of generality (resp. specificity) of a concept in terms of an information content measure that is based on concept refinement.
The measure is defined in detail in Section~\ref{sec:trepan_reloaded} 
and serves as the basis for the subsequent extension of the \trepan algorithm. 

\subsection{Concept refinement}

The idea behind concept refinement is to make a concept more general or more specific by means of refinement operators.  
Refinement operators are well-known in Inductive Logic Programming, where they are used to learn concepts from examples~\cite{vanderLaag98}. Refinement operators for description logics were introduced in~\cite{Lehmann2010}, and further developed in~\cite{Confalonieri2018,aaai2018}. In this setting, two types of refinement operators exist: 
specialisation refinement operators and generalisation refinement operators. 
While the former construct 
specialisations of hypotheses, the latter construct
generalisations.

In this paper we focus on specialisation operators. 
A specialisation operator takes a concept $C$ as input and returns a set of descriptions that are  more specific than $C$ by taking an ontology into account. The proposal laid out in this paper can make use of any such operators (see~e.g.,~\cite{Confalonieri2018,aaai2018,ijcai2018}). 
%
When a specific refinement operator is needed, as in the examples and in the experiments, we use the following definition of specialisation operator based on the {\em downcover} set of a concept $C$. 

\begin{definition}
Given a Tbox \Tmc, and a concept decription $C$, the specialisation operator $\rho_{\Tmc}(C)$ is defined as follows:
\begin{equation*}
\rho_{\Tmc}(C) \subseteq \mathsf{DownCov}_{\Tmc}(C) .
\end{equation*}
\noindent where $\mathsf{DownCov}_{\Tmc}(C)$ is the set of concepts that are more specific (or less general) than $C$:  
\begin{align}
\mathsf{DownCov}_{\Tmc}(C) := {} &\{ D \in \sub(\Tmc) \mid 
	D \sqsubseteq_{\Tmc} C  \text{ and}  \nonumber\\ &
	\nexists. D' \in \sub(\Tmc) \text{ with } D \sqsubset_{\Tmc} D' \sqsubset_{\Tmc} C \}. \nonumber
\end{align}
\end{definition}

\noindent In the above definition, $\sub(\Tmc)$ denotes the union of all the subconcepts in the axioms in $\Tmc$, plus $\{\top,\bot\}$. For any given axiom $C \sqsubseteq D$ in \Tmc,  
the set of its subconcepts is $\sub(C \sqsubseteq D) = \sub(C) \cup \sub(D)$; also, $\sub(C(a)) = \sub(C)$. Notice that $\sub(\Tmc)$ is a finite set. 

A concept $C$ is specialised by any of its most general specialisations that belong to $\sub(\Tmc)$. Every concept can be specialised into $\bot$ in a finite number of steps. 
\begin{definition}
The unbounded finite iteration  of the refinement operator $\rho$ is defined as:
\begin{equation*}
\rho_{\Tmc}^*(C) = \bigcup_{i \geq 0} \rho_{\Tmc}^i(C).
\end{equation*}
\noindent where $\rho_{\Tmc}^i(C)$ is inductively defined as:
\begin{align*}
\rho_{\Tmc}^0(C) &= \{C\},\\ \rho_{\Tmc}^{j+1}(C) &= \rho_{\Tmc}^j(C) \cup \bigcup_{C' \in \rho_{\Tmc}^j(C)} \rho_{\Tmc}(C'), j \geq 0.
\end{align*}
\end{definition}
Thus $\rho_{\Tmc}^*(C)$ is the set of subconcepts of $C$ w.r.t. \Tmc. We will denote this set by $\mathsf{subConcept}(C)$.
Since $\sub(\Tmc)$ is a finite set, the operator $\rho_{\Tmc}^*(C)$ is finite, and it terminates. For a detailed analysis of properties of refinement operators in DLs we refer to~\cite{Lehmann2010,Confalonieri2018}. 

\begin{example}\label{ex:example1}
Let us consider the concepts $\onto{Entity}$, and $\onto{LoanApplicant}$ defined in the ontology in Figure~\ref{fig:ontology}.  
Then: $\rho_{\Tmc}(\onto{Entity}) \subseteq \{\onto{Entity},$ $ \onto{AbstractObject}, \onto{PhysicalObject}$,
\onto{Quality}$\}$; 
$\rho_{\Tmc}^*(\onto{Entity}) \subseteq \mathsf{sub(\Tmc) \backslash \{\top\} }$;
$\rho_{\Tmc}(\onto{LoanApplicant})$ = $\rho_{\Tmc}^*(\onto{LoanApplicant}) \subseteq \{\onto{LoanApplicant}, \bot\}$.  
\end{example}

\section{\TREPAN RELOADED}
\label{sec:trepan_reloaded}

Our aim is to create decision trees that are more understandable for humans by determining which features are more understandable for a user, and assigning priority in the tree generation process according to increased understandability.
Our hypothesis, which we will validate in this paper, is that 
features are more understandable if they are associated to more general concepts present in an ontology. 

To measure the degree of semantic generality or specificity of a concept, we consider its \textit{information content}~\cite{Sanchez2011} as typically adopted in computational linguistics~\cite{Resnik1995}. 
There it quantifies 
the information provided by a concept when appearing in a context. 
Classical information theoretic approaches compute the information content of a concept as the inverse of its appearance probability in a corpus, so that
infrequent terms are considered more informative than frequent ones. 


In ontologies, the information content can be computed either extrinsically from the concept occurrences (e.g.,~\cite{Resnik1995}), or intrinsically, according to the number of subsumed concepts modeled in the ontology. 
Here, we adopt the latter approach. 
We use this degree of generality to prioritise features 
that are more general 
(thus presenting less information content), as our assumption is that the decision tree becomes more understandable when it uses more general concepts.
From a cognitive perspective this appears reasonable, since more general concepts have been found to be 
easier 
to understand and learn~\cite{Eleanor1976BasicOI}, 
and we test this assumption empirically below. 

\begin{definition} 
Given an ontology \Tmc, the information content of a feature $X_i$ is defined as:
\[
 \mathsf{IC}(X_i) := 
  \begin{cases} 
  1 - \cfrac{\log{(|\mathsf{subConcepts}(X_i)|)}}{\log{(|\mathsf{sub}(\Tmc)|)}} & \text{if } X_i \in \mathsf{sub}(\Tmc) \\
   0  & \text{otherwise.}
  \end{cases}
\]
\end{definition}

\noindent where $\mathsf{subConcepts}(X_i)$ is the set of specialisations for $X_i$, and $\sub(\Tmc)$ is the set of subconcepts that can be built from the axioms in the TBox \Tmc of the ontology (see~Section~\ref{sec:ontologies}). 

It can readily be seen that the values of $\mathsf{IC}$ are smaller for features associated to more general concepts, and larger for those associated to more specific concepts instead.
\begin{example}
Let us consider the concepts $\onto{Entity}$, and $\onto{LoanApplicant}$ defined in the ontology in Figure~\ref{fig:ontology} and the refinements in Example~\ref{ex:example1}. The cardinality of $\mathsf{sub}(\Tmc)$ is $13$. The cardinality of  $\mathsf{subConcepts}(\onto{Entity})$ and $\mathsf{subConcepts}(\onto{LoanApplicant})$ is 12 and 2 respectively. Then: $\mathsf{IC}(\onto{Entity})=0.04$, 
and $\mathsf{IC}(\onto{LoanApplicant})=0.73$.
\end{example}

\noindent Having a way to compute the information content of a feature $X_i$, 
we now propose to update the information gain used by \trepan to give preference to features with a lower information content. 
\begin{definition}
The information gain given the information content $\mathsf{IC}$ of a feature $X_i$ is defined as:
\[
 \mathsf{IG'}(X_i,S | \mathsf{IC}) := 
  \begin{cases} 
   (1-\mathsf{IC}(X_i)) \mathsf{IG}(X_i,S)  & \text{if } 0 < \mathsf{IC}(X_i) < 1 \\ 
   0  & \text{otherwise.}
  \end{cases}
\]
\noindent where $\mathsf{IG}(X_i,S)$ is the information gain as usually defined in the decision tree literature. 
\end{definition}
According to the above equation, $\mathsf{IG'}$ of a feature is decreased by a certain 
proportion %
that varies depending on its information content, and is set to $0$ either when the feature is not present in the ontology or when its information content is maximal.

Our assumption that using features associated with more general concepts 
in the creation of split nodes can enhance the understandability of the tree, is based on users  being more familiar with more general concepts rather than 
more specialised ones. 
To validate this hypothesis we ran a survey-based online study with human participants. 
Before proceeding to the details of the study and the results, 
as a prerequisite we introduce two measures for the understandability of a decision tree---an {\em objective}, syntax-based and a {\em subjective}, performance-based one---in the following section. 

\section{UNDERSTANDABILITY OF DECISION TREES}
\label{sec:understandability}

\emph{Understandability} depends not only on the characteristics of the tree itself, but also on the cognitive load experienced by users in using the decision model to classify instances, and in understanding the features in the model itself. 
However, for practical processing, understandability of decision trees needs to be approximated by an objective measure.  
We compare here two characterisations of the understandability of decision trees, approaching the topic from these two different perspectives: 
\begin{itemize}
    \item Understandability based on the syntactic complexity of a decision tree.
    \item Understandability based on users' performances, reflecting the cognitive load in carrying out tasks using a decision tree.
\end{itemize}
\noindent On the one hand, it is desirable to provide a technical characterisation of understandability that can give a certain control over the process of generating explanations. For instance, in \trepan, experts might want to stop the extraction of decision trees that do not overcome a given tree size limit, do have a stable accuracy/fidelity, but have an increasing syntactic complexity.

Previous work attempting to measure the understandability of symbolic decision models~(e.g.,~\cite{Huysmans2011}), and decision trees in particular~\cite{Piltaver2016}, proposed syntactic complexity measures based on the tree structure. The syntactic complexity of a decision tree can be measured, for instance, by counting the number of internal nodes in the tree or leaves, the number of symbols used in the splits (relevant especially for {\em m-of-n} splits), or the number of branches that decision nodes have.

For the sake of 
simplicity, we focus  
on the combination of two syntactic measures: 
the number of leaves $n$ in a decision tree, and the number of branches $b$ on the paths from the root of the tree to all the leaves in the decision tree. 
Based on the results in~\cite{Piltaver2016}, we define the {\em syntactic complexity} of a decision tree as:
\begin{align}\label{eq:syntactic_complexity}
U(n,b) := \alpha \cfrac{n}{k} + (1-\alpha) \cfrac{b}{k^2}   .
\end{align}
\noindent with $\alpha \in [0,1]$ being a tuning factor that adjusts the weight of $n$ and $b$, and $k = 5$ being the coefficient of the linear regression built using the results in~\cite{Piltaver2016}. 

On the other hand, the syntactic complexity of decision trees does not necessarily capture the ease with which actual people can use the resulting trees. A direct measure of user understandability is how accurately a user can employ a given decision tree to perform a decision. An often more precise measure of cognitive difficulty in mental processing is the reaction time (RT) or response latency~\cite{Donders1969OnTS}. RT is a standard measure used by cognitive psychologists and has even become a staple measure of complexity in the domain of design and user interfaces~\cite{principles03}. In the following section we describe an experiment measuring the cost of processing in terms of accuracy, and RT (among other variables) for different types of decision trees.

An additional factor that has to be taken into account is the tree size. It seems very likely that  trees of different sizes, irrespective of any actual complexity, present more difficulties for human understanding that are not necessarily linearly related to the increase in tree size. Therefore,  properly understanding the effects on actual understandability requires explicitly controlling the tree sizes. For our experiments, we define three categories of tree sizes based on the number of internal nodes: {\em small} (the number of internal nodes is between $0$ and $10$), {\em medium} (the number of internal nodes is between $11$ and $20$), and {\em large} (the number of internal nodes is between $21$ and $30$).

\section{EXPERIMENTAL EVALUATION}
\label{sec:evaluation}

\subsection{Methods}

\begin{figure*}[t!]
    \centering
    \subfloat[]
    {
        \includegraphics[width=.4\linewidth]{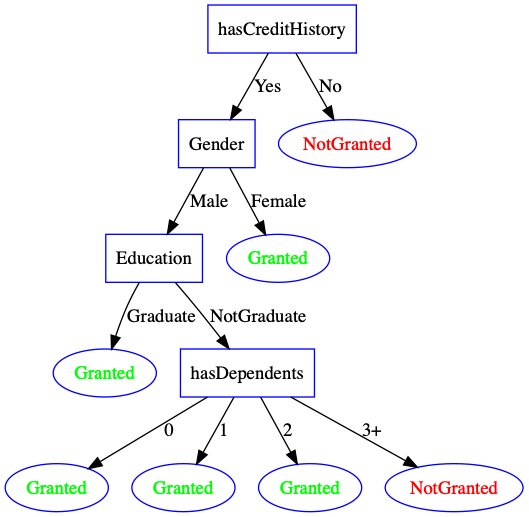}
    }
    ~
    \subfloat[]
    {
        \includegraphics[width=.4\linewidth]{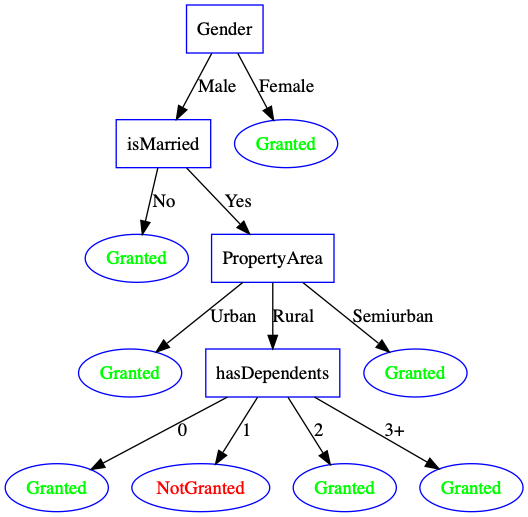}
    }
    \caption{Decision trees of size `small' in the loan domain, extracted without (left) and with (right) a domain ontology.
    As it can be seen the features used in the creation of the conditions in the split nodes are different.}
    \label{fig:decision_tree}
\end{figure*}

\paragraph{Materials.}
We used datasets from two different domains to evaluate our approach: finance and medicine. 
We used the 
Cleveland Heart Disease Data Set 
from the UCI archive\footnote{ \texttt{http://archive.ics.uci.edu/ml/datasets/Heart+Disease}}, and a loan dataset from Kaggle\footnote{\url{https://www.kaggle.com/altruistdelhite04/loan-prediction-problem-dataset}}. 
For each of them, we developed an ontology defining the main concepts and relevant relations (the heart and loan ontology contained 29 classes, 66 logical axioms and 28 classes, 65 logical axioms respectively). 
To extract decision trees using the \trepan and \trepan Reloaded algorithm, we trained two artificial neural networks implemented in {\em pytorch}. The neural networks we use in our experiments have a single layer of hidden units. The number of hidden units used for each network is chosen using cross-validation on the network's training set, and we use a validation set to decide when to stop training networks. The accuracy of the trained neural networks was of $85.98\%$ and $94.65\%$ for the loan and heart dataset respectively. 
In total, for each of the neural networks, we constructed six decision trees, varying their size (measured in number of nodes; i.e., small, medium, large), and whether or not an ontology had been used in generating them. 
In this manner, we obtained a total of twelve distinct decision trees (2 domains $\times$ 3 sizes $\times$ 2 ontology presence values). 
Figure~\ref{fig:decision_tree} shows two examples of distilled decision trees. 
The (avg.) fidelity of the extracted trees was of $92.73\%$ (\trepan) $92.63\%$ (\trepan Reloaded) and $89.23\%$ (\trepan) $88.17\%$ (\trepan Reloaded) for the loan and heart dataset respectively (see~also~Table~\ref{tbl:accuracy_fidelity}). Notice that since the trees are post-hoc explanations of the artificial neural network, the fidelities of the distilled trees, rather than their accuracies, are the crucial measure. 

\paragraph{Procedure.}
The experiment used two online questionnaires on the usage of decision trees.
The questionnaires contained an introductory and an experimental phase.

In the introductory phase, subjects were shown a short 
video about decision trees, and how they are used 
for classification. 
In this phase, participants were asked to provide information on their age, gender, 
education, and on their 
familiarity with decision trees. 
    
The experiment phase was subdivided into 
two tasks: classification, and inspection. 
Each task starts with an instruction page describing the task to be performed. 
In these 
tasks the participants were presented with the six trees corresponding to one of the two domains. 
In the classification task, subjects were asked to use a decision tree to 
assign one of two classes to %
a given 
case %
whose features are reported in a table (e.g., \textit{Will the bank grant a loan to a male person, with 2 children, and a yearly income greater than \euro 50.000,00?}). 
In the inspection task, participants had to decide on the truth value of a particular statement  (e.g., \textit{You are a male; your level of education affects your eligibility for a loan.}). 
The main difference between the two types of questions used in the two tasks is that the former provides all details necessary for performing the decision, whereas the latter only specifies whether a subset of the features influence the decision. 
In these two tasks, for each tree, we recorded:
\begin{itemize}
\item Correctness of the response.
\item Confidence in the response, as provided on a scale from $1$ to $5$ (`Totally not confident'=1, \dots, `Very confident'=5).
\item Response time measured from the moment the tree was presented.
\item Perceived tree understandability as provided on a scale from $1$ to $5$ ( `Very difficult to understand'=1, \dots, `Very easily understandable'=5).
\end{itemize}


\paragraph{Participants.}
63 participants (46 females, 17 males) volunteered to take part in the experiment via an online survey.\footnote{The participants were recruited among friends and acquaintances of the authors.} Of these 34 were exposed to trees from the finance domain, and 29 to those in the medical domain. The average age of the 
participants is 33 ($\pm$ 12.23) years (range: 19 -- 67). In terms of educational level their highest level was a Ph.D. for 28 of them, a Master degree for 9 of them, a Bachelor for 12, and a high school diploma for 14. 47 of the respondents reported some familiarity with the notion of decision trees, while 16 reported no such familiarity. 

\subsection{Results}

We fitted a mixed-effects logistic regression model~\cite{BAAYEN2008390} predicting the correctness of the responses in the classification and inspection tasks. The independent fixed-effect predictors were the syntactic complexity of the tree, 
the presence or absence of an ontology in the tree generation, the task identity (classification vs. inspection), and the domain (financial vs. medical), as well as all possible interactions between them, as well as a random effect of the identity of the participant. 
\begin{figure}[t!]
    \centering
    \subfloat[]
    {
        \includegraphics[width=0.75\columnwidth]{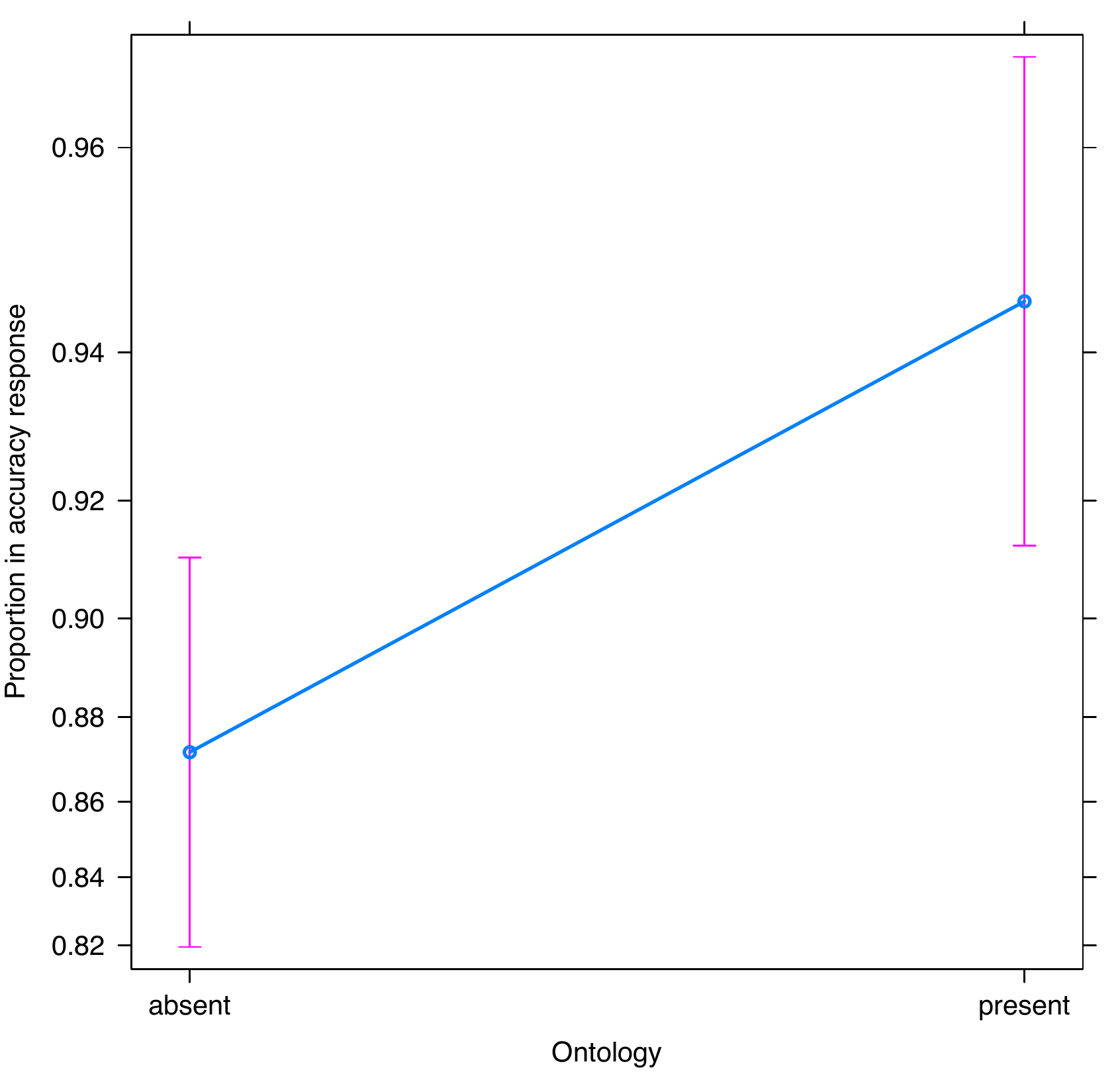}
    }
    \qquad
    \subfloat[]
    {
        \includegraphics[width=0.75\columnwidth]{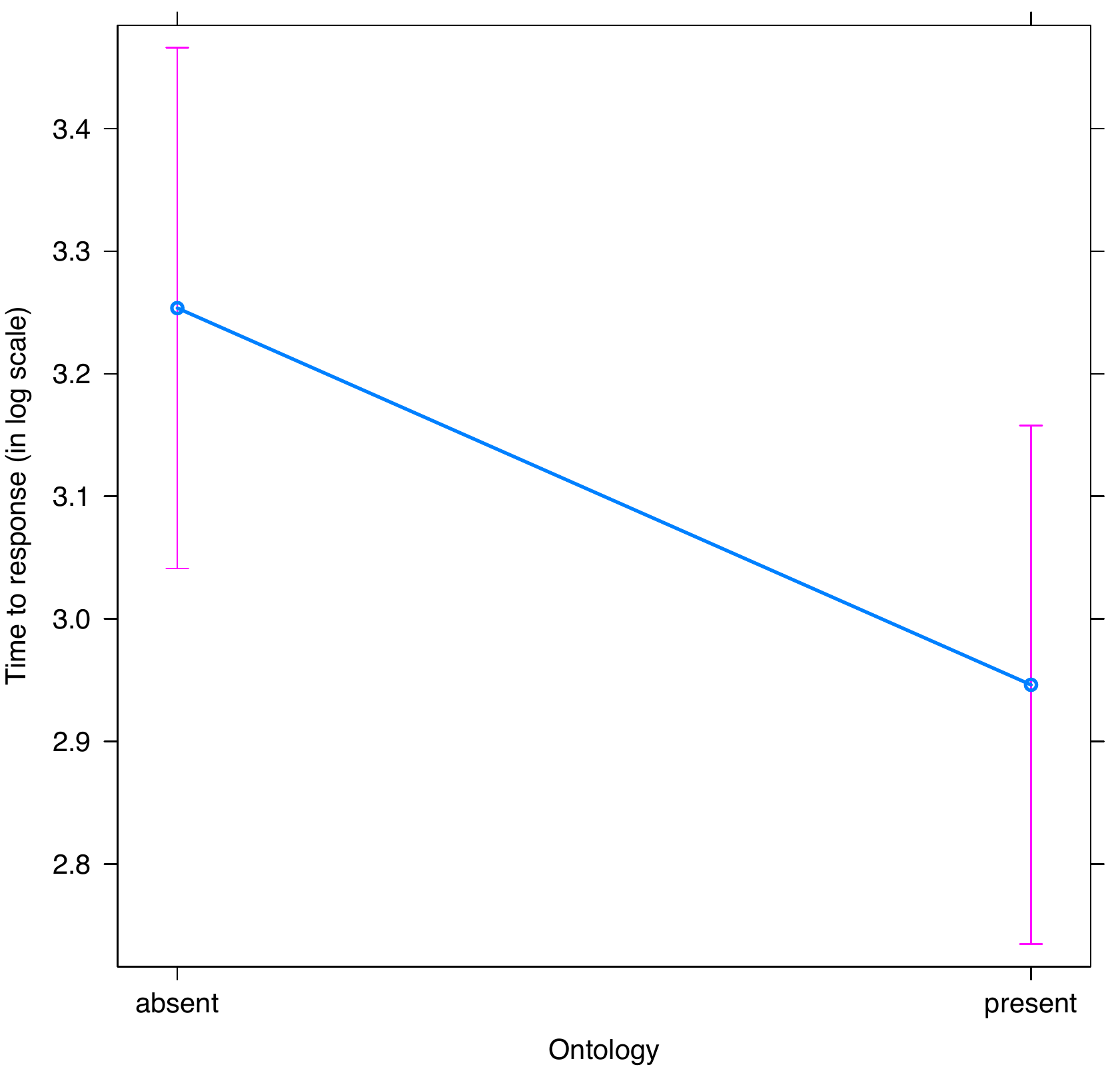}
    }
    \caption{Estimated main effects of ontology presence on accuracies (top) and time of response (bottom).}
    \label{fig:ontology_effects}
\end{figure}
%
%
A backwards elimination of factors revealed significant main effects of the task identity,  indicating that responses were more accurate in the classification task than they were in the inspection ($z=-3.00, p=.0027$), of the syntactic complexity ($z=-3.47, p=.0005$), by which more complex tree produced less accurate responses, and of the presence of the ontology ($z=3.70, p=.0002$), indicating that trees generated using the ontology indeed produced more accurate responses (Figure~\ref{fig:ontology_effects}a).
We did not observe any significant interactions or effect of the domain identity. 

We analysed the response times (on the correct responses) using a linear mixed-effect regression model~\cite{BAAYEN2008390}, with the log response time as the independent variable. As before, we included as possible fixed effects the task identity (classification vs inspection), the domain (medical vs financial), the syntactic complexity of the tree, and the presence or absence of ontology in the trees' generation, as well as all possible interactions between them. In addition, we also included the identity of the participant as a random effect. A step-wise elimination of factors revealed main effects of task identity ($F(1,593.87)=20.81, p<.0001$), syntactic complexity ($F(1,594.51)=92.42, p<.0001$), ontology presence ($F(1,594.95)=51.75,p<.0001$), as well as significant interactions between task identity and syntactic complexity ($F(1,594.24)=4.06, p=.0044$), and task identity and domain ($F(2,107.48)=5.03,p=.0008$). 
In line with what we observed in the accuracy analysis, we find that those trees that were generated using an ontology were processed faster than those that were generated without one (see Figure~\ref{fig:ontology_effects}b).

We analysed the user confidence ratings using a linear mixed-effect regression model, with the confidence rating as the independent variable. We included as possible fixed effects the task identity (classification vs inspection), the domain (medical vs financial), the size of the tree, and the presence or absence of ontology in the trees' generation, as well as all possible interactions between then. In addition, we also included the identity of the participant as a random effect. A stepwise elimination of factors revealed  a main effect of ontology presence ($F(1,689)=14.38, p=.0002$), as well as significant interactions between task identity and syntactic complexity ($F(2,689)=46.39, p<.0001$), and task identity and domain ($F(2,110.67)=3.11,p=.0484$). These results are almost identical to what was observed in the response time analysis: users show more confidence on judgments performed on trees that involved an ontology, the effect of syntactic complexity is most marked in the inspection task, and the difference between domains only affects the classification task.

Finally, we also analysed the user rated understandability ratings using a linear mixed-effect regression model, with the confidence rating as the independent variable. We included as possible fixed effects the task identity (classification vs inspection), the domain (medical vs financial), the syntactic complexity of the tree, and the presence or absence of ontology in the trees' generation, as well as all possible interactions between then, and an additional random effect of the identity of the participant as a random effect. A stepwise elimination of factors revealed significant main effects of task ($F(1,690)=27.21,p<.0001$), syntactic complexity ($F(1,690)=104.67,p<.0001$), and of the presence of an ontology ($F(1,690)=39.90, p<.0001$). These results are in all relevant aspects almost identical to what was observed in the accuracy analysis: the inspection task is harder, more syntactically complex trees are less understandable than less complex ones, and trees originating from an ontology are perceived as  more understandable.

\section{DISCUSSION}\label{sec:discussion}

Our hypothesis was that the use of ontologies to select features
for conditions in split nodes, as described above, leads to decision trees that are easier to understand. 
This ease of understanding was measured theoretically using a syntactic complexity measure, and cognitively through time and accuracy of responses as well as reported user confidence and understandability. 

First of all, the syntactic complexity (Eq.~\ref{eq:syntactic_complexity}) of the trees distilled with \trepan Reloaded is slightly smaller than those generated with \trepan (see Table~\ref{tbl:syntactic_complexity}). Such small reduction on syntactic complexity might or might not reflect differences in the actual understandability of the distilled trees by people. However, in our experiments, all online implicit measures (accuracy and response time), and off-line explicit measures (user confidence and understandability ratings) indicate that trees generated using an ontology are significantly more accurately and easier understood by people than are trees generated without ontology. The analyses of the four measures are remarkably consistent in this crucial aspect~(see~Table~\ref{tbl:results}). 
\begin{table}
\caption{Syntactic complexity (Eq.~\ref{eq:syntactic_complexity}) for trees inferred using C4.5, and distilled using \trepan, and \trepan Reloaded respectively.}
\begin{tabular}{cccc}
\hline 
 & C4.5   & \trepan & \trepan Reloaded \\
\hline \hline
heart   & 5.64  &  3.56  & {\bf 3.46} \\
loan    & 5.9   &  2.89  & {\bf 2.63} \\
\end{tabular}
\label{tbl:syntactic_complexity}
\end{table}
\begin{table}
\caption{Mean values of correct answers, time of response, user confidence, and user understandability for trees distilled using \trepan and \trepan Reloaded (standard deviations are reported in paranthesis). The difference in results is statistically significant w.r.t. Mann-Whitney and Wilcoxon tests for all measures.
}
\begingroup
\setlength{\tabcolsep}{10pt} 
\begin{tabular}{llll}
    \hline
    Task & Measure & \trepan & \trepan~Reloaded \\
    \hline \hline
    \multirow{3}{*}{Class.} 
    & \%C.~Answers & 0.87 (0.32)  & {\bf 0.94} (0.18) \\
    & Time (sec) & 43.25 (61.16)  & {\bf 24.29} (15.67) \\
    & Confidence & 4.38 (0.86) & {\bf 4.56} (0.80) \\
    & Understd. & 4.06 (0.97) & {\bf 4.50} (0.48) \\
    \multirow{3}{*}{Insp.} & $~$ & $~$ & $~$ \\
    & \%C.~Answers & 0.78 (0.41) & {\bf 0.90} (0.27) \\
    & Time (sec) & 35.90 (24.80)  & {\bf 26.55} (35.74) \\
    & Confidence & 4.10 (0.96)  & {\bf 4.36} (0.78) \\
    & Understd. & 3.83 (1.01) & {\bf 4.20} (0.87) \\
\end{tabular}
\endgroup
\label{tbl:results}
\end{table}
\begin{table}
\caption{Test-set accuracies and fidelities for trees distilled using \trepan and \trepan Reloaded.}
\begingroup
\setlength{\tabcolsep}{3pt} 
\begin{tabular}{l|llll|ll}
\hline
 & \multicolumn{4}{c}{Accuracy} & \multicolumn{2}{c}{Fidelity} \\
\hline \hline
 &  C4.5    & NN  & \trepan & \trepan Rld. & \trepan & \trepan Rld. \\
\hline
heart   &  81.97\% & 94.65\% & 82.43\% & 80.87\% & 89.23\% & 88.17\%\\
loan    &  80.48\% & 85.98\% & 86.03\% & 82.80\% &  92.73\% & 92.63\% \\
\end{tabular}
\endgroup
\label{tbl:accuracy_fidelity}
\end{table}

As we anticipated, coercing the outputs of \trepan onto a pre-determined ontology (as in \trepan reloaded) impacts the fidelity (and accuracy) of the resulting trees (see Table~\ref{tbl:accuracy_fidelity}). Crucially, however, the very small compromise in the fidelity (on both examples, a drop of around one percent) of the neural network reconstruction is more than compensated for by the substantial improvement in the ease with which actual people can understand the resulting trees. When the goal is providing model explanations that are actually understandable by people, such a small compromise in fidelity is well worth it. Notice that, if we were not willing to compromise on fidelity at all, it would not make any sense to deviate in any amount from the original neural network's performance (i.e., any fidelity below 100\% would be unacceptable). In such case, however, one would retain the lack of user understandability of the models.

At this point, one might wonder why we should bother to create surrogate decision trees from black-box models, rather than inferring them directly from data. 
As already noticed in the original \trepan work~\cite{Craven1995}, distilling trees from networks can actually result in \emph{better} trees than those one would obtain by building the decision trees directly. To demonstrate this point, we also trained decision trees directly from the datasets using the classical C4.5 algorithm. Table~\ref{tbl:accuracy_fidelity} shows that the trees inferred by the \trepan variants are as accurate --if not more-- than those inferred directly. Moreover, the trees built directly had syntactic complexities that roughly doubled those of the trees distilled using either \trepan variant (see Table~\ref{tbl:syntactic_complexity}). This indicates that 
constructing trees directly from the data results in trees substantially more complex than those distilled by \trepan variants, that nevertheless do not outperform them in the task. 

There is a similarly small compromise in the accuracy of the decision trees (see~Table~\ref{tbl:accuracy_fidelity}). As we discussed above, in this approach, the accuracy of the resulting trees (i.e., their ability to replicate the testing sets) is less relevant than their fidelity (i.e., their ability to replicate the behaviour of the model we intend to explain). Nevertheless, our \trepan Reloaded method improves the understandability of the trees w.r.t. the original \trepan, while compromising little on the accuracy. 

Apart from improving the understandability of (distilled) decision trees, ontologies also pave the way towards the capability of changing the level of abstraction of explanations to match different user profiles or audiences. For instance, the level of technicality used in an explanation for a medical doctor should not be the same as that used for lay users. One wants to adapt explanations without changing the underlying explanation procedure. Ontologies are amenable to automated abstractions to improve understandability~\cite{Keet07}. The idea of concept refinement adopted here can be extended to operate on changing the definition of concepts and make them more general or more specific by means of refinement operators~\cite{Confalonieri2018,aaai2018,ijcai2018}. This is a line of work that we find a natural continuation of the current study.

In its current form, \trepan Reloaded requires a predefined ontology onto which the features used by our algorithm should be mapped. 
In such cases, which are common in many domains (e.g., medical, pharmaceutical, legal, biological, etc.), one can directly apply \trepan Reloaded to improve the quality of the explanations. Additional work --beyond the scope of the current study-- would be to automatically construct the most appropriate ontology to be mapped onto. Such a process could be achieved by automatically mapping sets of features into pre-existing general domain ontologies (e.g., MS Concept Graph~\cite{MSGraph12}, DBpedia~\cite{dbpedia-swj}). The provision of some form of explicit knowledge, rather than being particular to our method, resides at the core of any attempts at human interpretable explanations.
Whether such knowledge is in the form of a domain-specific ontology (as in this study), or as a domain-general one to be adapted ad-hoc, will depend on the particulars of specific applications.

\section{RELATED WORKS}
\label{sec:related_work}

Most approaches on interpretable machine learning focus either on building directly interpretable models, or 
on reconstructing {\em post-hoc} local explanations. 
Our approach belongs to the category of post-hoc global explanation methods. 


In this latter category, there are a few approaches that closely relate to ours. 
For instance, the work in \cite{Renard2019ConceptTree} uses concepts to 
group features (either using expert knowledge or correlations), and embed them into surrogate models in order to constrain their training. In particular, the authors propose a revised version of \trepan that considers only features belonging to concepts in the extraction of a decision tree. 
Whilst their results show that surrogate trees preserve accuracy and fidelity compared with original versions, the improvement in human-readability is not explicitly tested with users. 
The approach in~\cite{Dhurandhar2019ibm} uses a complex neural network model to improve the accuracy of a simpler model, e.g., a simpler neural network, or a decision tree. This approach assumes to have a white-box access to (some of) the layers of the complex network model, whereas, in our approach we treat the black-box as an oracle. The authors in~\cite{BastaniKB17} describe a method for extracting decision tree explanations that actively samples new training points to avoid overfitting. Our approach is similar since \trepan also uses new sampled data during the extraction of the decision tree.

Other works focus on building terminological decision trees from and using ontologies, e.g.,~\cite{RIZZO20171,ZhangSH02}). These approaches perform a classification task while building a tree rather than building a decision tree from a classification process computed by a black-box. 


\section{CONCLUSION AND FUTURE WORKS}
\label{sec:conclusion}

In this paper, we proposed an extension of \trepan, an algorithm that extracts global post-hoc explanations of black box-models in the form of decision trees. Our algorithm \trepan Reloaded takes into account ontologies in the distillation of decision trees. 

We showed that the use of ontologies ease the understanding of the distilled trees by actual users. We measured this ease of understanding through a rigorous experimental evaluation: theoretically, using a syntactic complexity measure, and, cognitively, through time and accuracy of responses as well as reported user confidence and understandability. All our measures indicated that trees distilled by \trepan Reloaded are significantly more accurately and easier understood by people than are trees generated by \trepan, with only little compromise of the accuracy 
and the fidelity 
(see~Section~\ref{sec:discussion}).  

The results obtained are very promising, and they open several direction of future research. On the one hand, we plan to extend this work to support the automatic generation of explanations that can accommodate different user profiles. 
On the other hand, we aim at investigating to apply our approach to explain CNNs in image classification (e.g.,~\cite{Zhang2019CNNdtrees}). We also believe that this approach can be useful in bias identification, to understand, for instance, if  any undesirable discrimination features are affecting a black-box model.


\bibliographystyle{ecai}
\bibliography{xai-bibliography,ecai2020}

\end{document}